\documentclass[letterpaper, 10 pt, conference]{ieeeconf}  
\usepackage{amsmath}  
\usepackage{graphicx}
\usepackage[caption=false,font=footnotesize]{subfig}  
\usepackage{xcolor}  
\usepackage{booktabs} 
\usepackage{bm}

\usepackage{booktabs}
\usepackage{siunitx}
\usepackage{pgfplotstable}
\usepackage{amssymb} 

\IEEEoverridecommandlockouts                              
\title{\LARGE \bf
Swimming Under Constraints: A Safe Reinforcement Learning Framework for Quadrupedal Bio-Inspired Propulsion
}

\author{
    Xinyu Cui$^{1,2,3  *}$, Fei Han$^{1,4  * \dagger}$, Hang Xu$^1$, Yongcheng Zeng$^{2,3}$, Luoyang Sun$^{2,3}$, Ruizhi Zhang$^2$, \\
    Jian Zhao$^5$, Haifeng Zhang$^2$, Weikun Li$^1$, Hao Chen$^1$, Jun Wang$^6$, Dixia Fan$^{1 \dagger}$ \\
    \thanks{* Equal contribution, the order is determined by flipping a coin.}
    \thanks{$\dagger$ Corresponding to: \{hanfei, fandixia\}@westlake.edu.cn}
    \thanks{This work is supported by the Key Research and Development Program of Zhejiang Province (Grant No. 2023C03133), the Young Scientists Fund of the National Natural Science Foundation of China (Grant No. 52401393), the Natural Science Foundation of Hangzhou, Zhejiang Province, China (Grant No. 2025SZRJJ1217), and the Hangzhou Postdoctoral Daily Expenses Funding (Grant No. 103120046582502).}
    \thanks{$^{1}$ FSI lab, Westlake University, Hangzhou, 310030, China.}
    \thanks{$^{2}$ Institute of Automation, Chinese Academy of Sciences, Beijing, 100190, China.}
    \thanks{$^{3}$ School of Artificial Intelligence, University of Chinese Academy of Sciences, Beijing 100049, China}
    \thanks{$^{4}$ Zhejiang University, Hangzhou, 310027, China.}
    \thanks{$^{5}$ Zhongguancun Academy, Beijing, 100094, China}
    \thanks{$^{6}$ Department of Computer Science, University College London, London WC1E 6BT, United Kingdom}
}

\begin{document}

\maketitle
\thispagestyle{empty}
\pagestyle{empty}


\begin{abstract}
Bio-inspired aquatic propulsion offers high thrust and maneuverability but is prone to destabilizing forces such as lift fluctuations, which are further amplified by six-degree-of-freedom (6-DoF) fluid coupling. We formulate quadrupedal swimming as a constrained optimization problem that maximizes forward thrust while minimizing destabilizing fluctuations. 
Our proposed framework, Accelerated Constrained Proximal Policy Optimization with a PID-regulated Lagrange multiplier (ACPPO-PID), enforces constraints with a PID‑regulated Lagrange multiplier, accelerates learning via conditional asymmetric clipping, and stabilizes updates through cycle‑wise geometric aggregation.
Initialized with imitation learning and refined through on-hardware towing-tank experiments, ACPPO-PID produces control policies that transfer effectively to quadrupedal free-swimming trials. Results demonstrate improved thrust efficiency, reduced destabilizing forces, and faster convergence compared with state-of-the-art baselines, underscoring the importance of constraint-aware safe RL for robust and generalizable bio-inspired locomotion in complex fluid environments.
\end{abstract}


\section{Introduction}

Bio-inspired aquatic propulsion systems, such as flapping foils, undulating fins, and webbed paddles have attracted significant attention for their ability to generate large instantaneous thrust\cite{wang2024learn, triantafyllou2005review}, achieve high propulsive efficiency, and enable agile maneuverability \cite{xia2024combined, jung2023bioinspired, li2021design}. However, unlike conventional screw or jet drives, the asymmetric morphology and unsteady interactions of these propulsors often introduce destabilizing effects, including lift fluctuations, pitch oscillations, and lateral forces at peak thrust \cite{ priovolos2018vortex, barrett1999drag}. Such disturbances elevate energy consumption, compromise stability, and degrade mission performance. 

This challenge extends to all fluid-coupled vehicles operating underwater, on surface, or in air. Large pitch and roll excursions, with disturbances induced by wave or vortex, increase drag and impair maneuverability \cite{fossen2011handbook, colgate2004mechanics, picardi2023underwater}. Unsteady fluid–structure interactions also generate destabilizing moments that reduce efficiency and control authority \cite{etkin1995dynamics}. Therefore, locomotion design needs to constrain unwanted oscillatory forces while maximizing thrust.

While pre-defined gaits with fixed parameters can suppress some of these effects, they under-utilize the nonlinear dynamics of unsteady hydrodynamics and fail to fully leverage the robot’s embodiment \cite{han2025learn, ijspeert2008central, tong2022cpg}.
Reinforcement learning (RL) provides a promising alternative by enabling propulsors to autonomously discover control policies that exploit embodiment and hydrodynamics. However, naive RL exploration typically results in instability and inefficient convergence when balancing the dual objectives of maximizing thrust while minimizing unwanted forces\cite{kim2024wing}. Safe RL offers a principled framework to address this challenge: explicit constraints guide policy updates to ensure high performance without violating safety or stability bounds\cite{ray2019benchmarking}. 
While safe RL has seen success in ground and aerial robotics, its application to aquatic learning remains largely unexplored, where strong fluid coupling and costly experimentation make stability guarantees especially critical.

This paper uses fast forward swimming in a bio-inspired quadrupedal robot to validate our proposed safe reinforcement learning algorithm. We formulate gait learning as a constrained optimization problem that maximizes thrust while bounding lift oscillations—the primary destabilizing factor. To solve this, we propose Accelerated Constrained PPO with a PID-regulated Lagrange multiplier (\textbf{ACPPO-PID}). Our approach enhances on-hardware learning by dynamically enlarging the PPO \cite{schulman2017proximal} clip range under constraint satisfaction, and employing cycle-wise geometric aggregation for stability. The policy is seeded via Imitation Learning, refined through real-time towing-tank reinforcement learning, and deployed on a free-swimming robot using diagonal-phase paddling. Experiments show ACPPO-PID significantly outperforms state-of-the-art baselines in thrust efficiency, lift suppression, and convergence speed.

\begin{figure*}[t]
    \centering
    \includegraphics[width=0.9\linewidth]{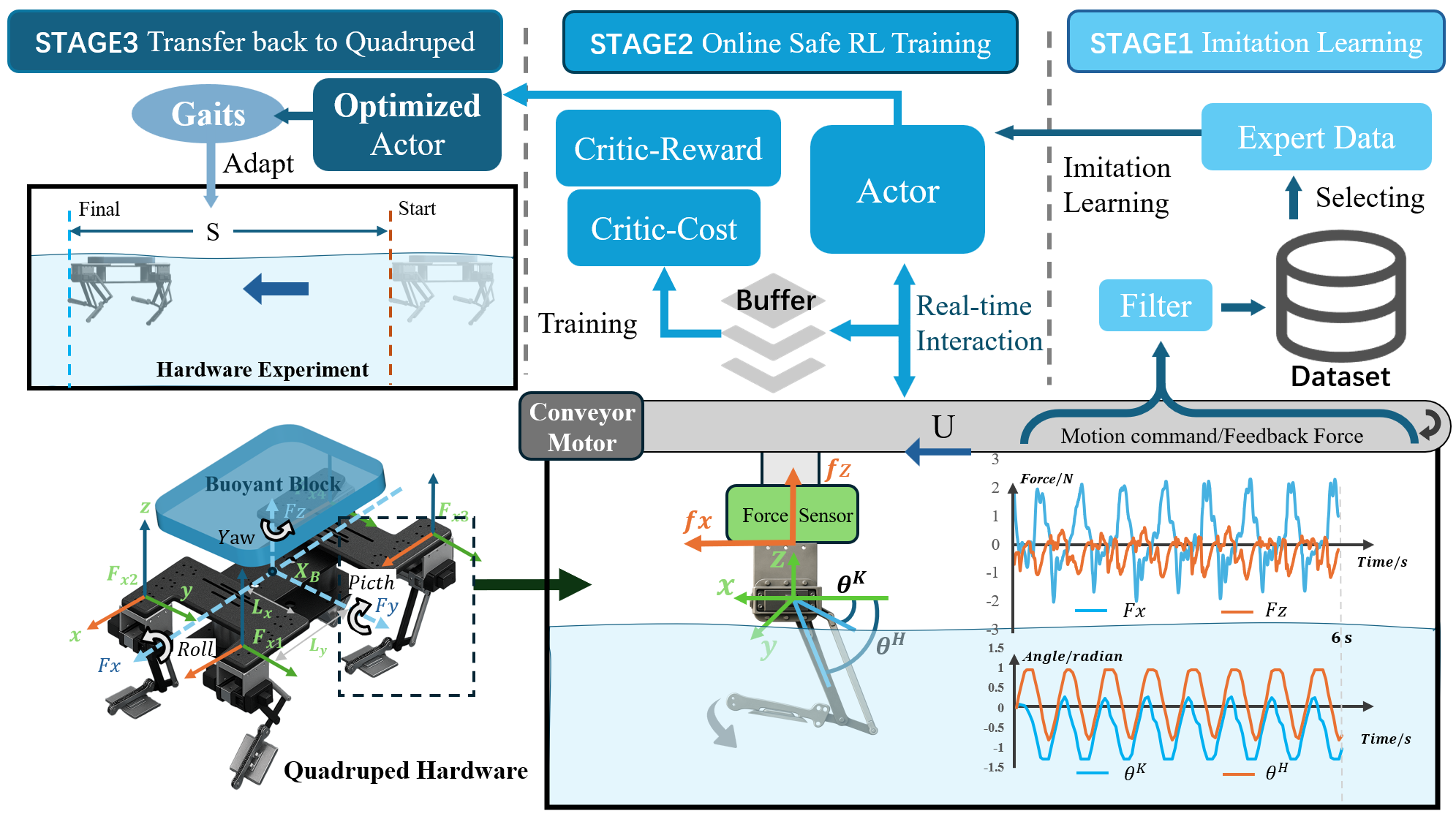} 
    \caption{Proposed framework: imitation learning initializes a periodic gait from predefined motions while sensor feedback is Kalman‑filtered; safe RL fine‑tuning accelerates on‑hardware convergence under stability constraints; and the resulting one‑cycle paddle is transferred to diagonal limb pairs with a half‑cycle phase offset to enable smooth and stable free‑swimming.}
    \label{workflow}
    \vspace{-0.15in}
\end{figure*}
\textbf{Contributions.} This paper makes three main contributions:
\begin{itemize}
\item We formulate quadrupedal swimming as a constrained thrust optimization problem, decoupling full-robot optimization into single-limb gait optimization.
\item We propose ACPPO-PID, a safe RL framework that balances broad exploration with strict constraint enforcement, accelerating on-hardware learning.
\item We validate the approach through real-world towing-tank and quadruped swimming experiments, demonstrating improved thrust, stability, and convergence over state-of-the-art baselines.
\end{itemize}

\section{Related Work}

\subsection{RL for Bio-inspired Robotic Locomotion}

RL has emerged as a powerful tool for optimizing locomotion in bio-inspired robots
In terrestrial settings, RL has been successfully applied to legged robots for robust walking and running \cite{peng2020learning, choi2023learning}, often learning directly from proprioceptive sensing and reward signals related to forward progress and energy efficiency. 
In aquatic environments, RL has been employed to optimize swimming gaits in robotic fish \cite{wang2024learn, cui2024enhancing}, undulating fins \cite{li2021design}, and other bio-inspired propulsors \cite{colgate2004mechanics}. 

However, these methods typically focus on maximizing thrust or speed without explicitly accounting for stability constraints or unwanted forces like lift in a forward-moving task. Additionally, their reliance on simulation may not accurately capture complex fluid-structure interactions, creating sim-to-real gaps. In contrast, our work emphasizes on-hardware learning with explicit stability constraints, directly addressing the reality gap and improving policy practicality.

\subsection{Safe RL for Constrained Optimization}

Safe RL addresses constrained optimization through several paradigms. Local policy search methods like CPO \cite{achiam2017constrained} extend TRPO \cite{schulman2015trust} with second-order approximations, while FOCOPS \cite{zhang2020first} reduces computational overhead through feasible-trust region intersections. Projection-based approaches such as PCPO \cite{yang2020projection} and CUP \cite{yang2022cup} optimize within trust regions before projecting policies into feasible spaces, often yielding conservative outcomes.

Primal-dual methods transform constraints into unconstrained Lagrangian forms, with RCPO \cite{tessler2018reward} adjusting multipliers for stability and CPPO-PID \cite{stooke2020responsive} incorporating PID controllers \cite{willis1999proportional} for improved convergence. While primal-dual approaches offer computational efficiency suitable for on-hardware learning, they exhibit performance instability and require careful hyperparameter tuning.
Our method addresses these limitations by enhancing exploration when safe while improving training stability through cycle-wise aggregation. This approach can also steer policies toward constraint satisfaction when violations persist, reducing hyperparameter sensitivity for practical deployment.

\section{Preliminaries}

To validate the optimization framework and coordination strategy, we developed two complementary systems: a \textbf{single-leg towing platform} for learning physical constraints in controlled flows, and a \textbf{quadrupedal underwater robot} for validating gait generalization in free-swimming scenarios. This setup forms a closed loop from isolated actuator learning to embodied whole-robot validation.

\subsection{Quadrupedal Underwater Robot and System Dynamics}

As illustrated in the schematic diagram of the robotic system (Fig.~\ref{workflow}, bottom left), the quadrupedal underwater robot comprises a support plate, four independent single-leg actuators, and modular buoyancy units, with overall dimensions of $480\,\text{mm} \times 250\,\text{mm} \times 88\,\text{mm}$. 

The dynamic model of the robot is derived based on its three-dimensional geometry, incorporating both the thrust forces and intrinsic joint torques generated by each leg actuator. Within this model, $X_B$ denotes the center of buoyancy. The position of each actuator is defined in a body-fixed coordinate system with origin at $X_B$, represented as $(L_x, L_y, h)$, where $L_x$ and $L_y$ are the horizontal offsets from $X_B$, and $h$ indicates the vertical eccentricity relative to the central rotational axis. For gait synthesis and analysis, the four foot actuators are abstracted as two diagonal pairs: $\{P_1, P_4\}$ and $\{P_2, P_3\}$, simplifying the control strategy while preserving dynamic characteristics.

\textbf{Diagonal coordination and constrained optimization: } Owing to the approximate geometric symmetry of the four limbs about both longitudinal and lateral axes, a diagonally symmetric coordination strategy is adopted. The four limbs are grouped into two diagonal pairs, and each pair is driven either in phase or with a half-cycle offset depending on the gait design. This strategy provides two benefits: (i) in-plane force components tend to cancel out in yaw due to symmetry, and (ii) the overlap of thrust peaks and troughs between diagonals smooths the net force profile, thereby reducing oscillations in propulsion and suppressing attitude disturbances. Since the limb kinematics are mechanically confined to the sagittal plane, lateral hydrodynamic forces are inherently minimized, with any residuals effectively neutralized by the symmetric coordination. As the limbs generate the primary propulsion and their motion patterns are assumed similar, the gait optimization problem of the quadruped can thus be reduced to optimizing a single representative limb, which is subsequently deployed to the whole robot via diagonal coordination.

In the validation experiments, the same single-leg motion strategy is mapped to diagonal pairs $\{P_1, P_4\}$ and $\{P_2, P_3\}$. Due to diagonal symmetry, the legs within each pair maintain synchronized motion and generate identical forces and moments. Let $\mathbf{F}_1$ and $\boldsymbol{\tau}_1$ represent the force and moment contribution from a single leg in the first pair, and $\mathbf{F}_2, \boldsymbol{\tau}_2$ for the second pair. With a vertical offset $h$ relative to the center of buoyancy, the net force and moments are derived as $\mathbf{F}_{\mathrm{total}}=2(\mathbf{F}_1+\mathbf{F}_2)=(f_X, f_Y, f_Z)$, $M_X=2(\tau_{1x}+\tau_{2x})-h f_Y$, $M_Y=2(\tau_{1y}+\tau_{2y})+h f_X$, and $M_Z=2(\tau_{1z}+\tau_{2z})$. This formulation demonstrates that yaw moments from in-plane forces cancel under diagonal symmetry, while the pitch and roll moments are directly coupled with the total horizontal thrust and the offset $h$.

\subsection{Two-DOF Single-Limb Platform and Learning Objective}

As shown in Fig.~\ref{workflow}(bottom right), a single-limb hardware platform with two degrees of freedom (DoF) is designed for controlled evaluation. The limb consists of a hip flexion/extension (HFE) joint and a knee flexion/extension (KFE) joint, mechanically coupled through a four-bar linkage. This configuration yields a compact structure while allowing large-amplitude, coordinated swinging motions. Each joint is independently actuated by a servo motor, and the motion is transmitted via rigid links to a web. A six-axis force/torque sensor is mounted near the web tip to measure instantaneous three-dimensional force and torque components, with emphasis on forward thrust \(F_x\), vertical lift \(F_z\), and pitching moment \(M_y\).

\textbf{Experimental setup: } The platform is tested in a towing tank, where the limb is mounted on a horizontal carriage translated at constant speed and fixed immersion depth while executing prescribed joint trajectories. Given commanded joint angles \(\theta^H\) and \(\theta^K\), the four-bar mechanism drives the web through a full oscillation cycle. During each cycle, the force/torque sensor synchronously records the time histories of \(F_x\), \(F_z\), and the associated moment components. This setup provides a repeatable and high-resolution means of evaluating hydrodynamic performance under different motion policies.

\textbf{Learning objective: } 
Based on the force-motion characteristics of the symmetrically arranged four legs of the quadruped robot platform, during diagonal-symmetric motion, the yaw moment \(M_Z\) and roll moment \(M_X\) are essentially zero, while the pitch moment remains consistently positive. As a result, in an unconstrained scenario, the robot moves in a straight line with a 'pitch-up' posture. Its motion effect is mainly influenced by the forward thrust \(f_X\) and the vertical force \(f_Z\), which causes vertical oscillations.
The control objective is formulated as maximizing the time-averaged forward thrust \(F_x\) while simultaneously suppressing both the magnitude and variability of the lift \(F_z\). This design explicitly reduces ineffective oscillatory work and improves locomotor stability. Crucially, the optimized single-limb policy can be directly transferred to the quadrupedal robot via diagonal coordination, thereby achieving high propulsion efficiency together with robust whole-body stability in free-swimming conditions.

\subsection{Problem Formulation}

Subsequently, we consider the optimization of a flapping limb to produce higher thrust while maintaining low oscillations in lift forces. To fully explore the hydrodynamic, we adopt a more agile end-to-end scheme instead of representing the limb's motion in a parametric sinusoidal form. We then formulate a reward function proportional to thrust while treating oscillation as a constraint.
Formally, we cast control of a single limb as a Constrained Markov Decision Process (CMDP) \cite{sutton1998introduction} $\langle \mathcal{S}, \mathcal{A}, P, R, C \rangle$, where:
\begin{itemize}
    \item State $s_t \in \mathcal{S}$: measured quantities of the limb, including joint phase, angular velocity, and the measured values of the sensors.
    \item Action $a_t \in \mathcal{A}$: the angle change of the joints in one control step.
    \item Transition $P(s_{t+1}\mid s_t,a_t)$: unknown dynamics, learned from interaction.
    \item Reward $R(s_t,a_t,s_{t+1})$: proportional to forward thrust component, denoted $r_t$.
    \item Cost $C(s_t,a_t,s_{t+1})$: lift non-cancellation across a half motion cycle, denoted $c_t$. With a cycle of $H$ steps, we define the instantaneous cost as $c_t=\bigl\lvert F_{z,t}+F_{z,t-H/2}\bigr\rvert$.
\end{itemize}

Let $\tau=(s_0,a_0,s_1,a_1,\dots)$ be a trajectory generated by policy $\pi(a\mid s)$ and discount factor $\gamma\in(0,1)$. The expected discounted return and cost are

\footnotesize 
\begin{equation}
\begin{aligned}
    J(\pi)&=\mathbb{E}_{\tau\sim\pi}\!\left[\sum_{t=0}^{\infty}\gamma^t\, R(s_t,a_t,s_{t+1})\right], \\
    J_C(\pi)&=\mathbb{E}_{\tau\sim\pi}\!\left[\sum_{t=0}^{\infty}\gamma^t\, C(s_t,a_t,s_{t+1})\right].
\end{aligned}
\end{equation}

\normalsize
The learning objective is

\footnotesize 
\begin{equation}
\begin{aligned}
\pi^{\star}\in\arg\max_{\pi}\ J(\pi)\quad \text{s.t.}\quad J_C(\pi)\le d,
\end{aligned}
\end{equation}
\normalsize
where $d>0$ bounds the allowable lift fluctuation. 

To solve this constrained optimization problem, we introduce the Lagrangian function, defined as

\footnotesize 
\begin{equation}
\begin{aligned}
&\quad \mathcal{L}(\pi, \lambda) = J(\pi) + \lambda (d - J_C(\pi)), \\
&= \mathbb{E}_{\tau\sim\pi}\!\left[\sum_{t=0}^{\infty}\gamma^t\, \left( R(s_t,a_t,s_{t+1}) - \lambda C(s_t,a_t,s_{t+1})\right) \right] + \lambda d, 
\end{aligned}
\end{equation}
\normalsize
where $\lambda \geq 0$ is the Lagrange multiplier. This formulation transforms the constrained problem into an equivalent unconstrained saddle-point optimization problem: $\min_{\lambda \geq 0} \max_{\pi} \mathcal{L}(\pi, \lambda)$.



\section{Methodology}
We propose a three-stage safe policy optimization framework, as illustrated in Fig.~\ref{workflow}. In the first stage, a periodic paddle motion is initialized through demonstration-driven imitation learning, providing a stable starting point for subsequent training. In the second stage, the policy is refined via on-hardware safe reinforcement learning in a towing tank, enabling constraint-aware adaptation under real hydrodynamic interactions. In the final stage, the optimized policy is transferred to the quadrupedal robot by recording a single cycle of joint trajectories and deploying it with diagonal-phase coordination, thereby achieving stable and efficient free-swimming in validation trials.

\subsection{Imitation Learning}

To reduce the wall-clock time required for on-hardware training, we first initialize the policy using \textbf{Imitation Learning (IL)} on curated demonstrations. These demonstrations are generated by parameterizing limb trajectories as sinusoids and performing a brute-force (BF) search over frequency, amplitude, and phase. The prescribed leg motion is expressed as:

\footnotesize 

\begin{equation}
\left\{
\begin{split}
    \theta^H(t) &= A_{\theta^H}  \sin(2\pi f t) + \theta^H_{0},\\
    \theta^K(t) &= A_{\theta^K}  \sin(2\pi f t + \phi) + \theta^K_{0},
\end{split}
\right.
\label{eq:brute_force}
\end{equation}
\normalsize
where, \(A_{\theta^H}\) and \(A_{\theta^K}\) represent the angles amplitude of the HFE joint and the KFE joint. The motion frequency is denoted by \(f\), and \(\phi\) is the phase difference between the HFE and KFE joints. \(\theta^H_{0}\) and \(\theta^K_{0}\) represent the initial offset in motion.

\begin{table}[htbp]
\caption{Experimental Parameter Ranges}
\vskip -0.25in
\begin{center}
\resizebox{\linewidth}{!}{
\begin{tabular}{|c|c|c|c|c|c|}
\hline
\textbf{Parameter}  & $A_{\theta^H}(rad)$ & $A_{\theta^K}(rad)$ & $f(Hz)$ & $\phi$ & $\theta^H_{0} / \theta^K_{0}$\\
\hline
\textbf{Range} & $[{\pi}/6, {\pi}/3]$ & $[{\pi}/12, {\pi}/4]$ & $[0.3, 0.6]$ & $[0, \pi]$ & $[{\pi}/4, {5\pi}/4]$\\
\hline
\end{tabular}
}
\label{tab:parameter_ranges1}
\end{center}
\vskip -0.15in
\end{table}
\normalsize


To generate expert demonstrations, we employ \textbf{Latin Hypercube Sampling (LHS)} over the parameter space in Eq.~\ref{eq:brute_force} and Table~\ref{tab:parameter_ranges1}, yielding over five thousand sinusoidal trajectories ranked by time‑averaged thrust, with the lowest‑lift subset retained as the demonstration set. Additionally, the best‑performing sinusoidal trajectory from the search is extracted and denoted as \textbf{BF}, which serves as the parameterized baseline in our experiments. Demonstration rollouts are stored as state–action pairs and are used to pretrain the policy via IL, enabling rapid convergence to a stable periodic gait. This provides a safe, efficient initialization for subsequent on‑hardware safe RL finetuning.


For both the policy and value functions, we adopt a \textbf{Transformer} architecture. Unlike recurrent models that compress history into a fixed hidden state which potentially losing high-frequency fluid details. Transformer's self-attention mechanism enables direct access to the entire observation window. This design is critical for aquatic propulsion, where hydrodynamic forces exhibit significant delays that recurrent bottlenecks often fail to capture. A sequence of recent observations is processed through the encoder, and the final embedding is decoded by a multilayer perceptron (MLP) to generate continuous actions, providing a robust foundation for stability-aware control.

\subsection{Safe RL with accelerated safe exploration}

We estimate the principal paddle frequency from $\mathcal{F}_z$, a sequence of $F_z$ observed in the replay buffer via a discrete Fourier transform (DFT). Let the control rate be $f_s=20\,\text{Hz}$. We remove low-frequency drift below $0.1\,\text{Hz}$ and restrict the search to $[0.1,5]\,\text{Hz}$ to reflect reciprocating paddles. The dominant frequency is

\footnotesize
\begin{equation}
\begin{aligned}
f^\star=\arg\max_{f\in[0.1,5]}\ \lvert \mathrm{DFT}(\mathcal{F}_z)\rvert,
\end{aligned}
\end{equation}
\normalsize
and the cycle (in steps) is $H=\left\lfloor \tfrac{f_s}{f^\star}\right\rfloor$. To promote lift cancellation under the half-cycle phase offset between limb groups, we have the instantaneous cost as
$c_t=\bigl\lvert F_{z,t}+F_{z,t-H/2}\bigr\rvert$, 
which directly penalizes residual lift after superposing signals separated by half a cycle.



To solve the constrained optimization problem introduced in the problem formulation, we employ an iterative algorithm that alternates between policy optimization and Lagrange multiplier updates. The process proceeds over iterations indexed by $k = 0, 1, 2, \dots$, starting with an initial $\lambda_0 \geq 0$ (typically set to 0).

At each iteration $k$, we first optimize the policy parameters $\theta$ to maximize the expected Lagrangian return using the current multiplier $\lambda_k$:

\footnotesize 
\begin{equation}
\begin{aligned}
\max_{\theta}\ \mathbb{E}\!\left[\sum_t \gamma^t \bigl(r_t - \lambda_k \, c_t\bigr)\right].
\end{aligned}
\end{equation}

\normalsize
Next, we collect trajectories under the updated policy $\pi_{\theta}$, compute an empirical estimate of the cost return $\hat{J}_C^{(k)}$, and measure the constraint violation $g_k = \hat{J}_C^{(k)} - d$.
Finally, we update the multiplier for the next iteration using PID control as defined in CPPO-PID~\cite{stooke2020responsive}:

\footnotesize 
$$\lambda_{k+1}=\Bigl[\lambda_k + K_P \, g_k + K_I \sum_{i=0}^{k} g_i + K_D (g_k - g_{k-1})\Bigr]_+,$$
\normalsize
where $[\cdot]_+$ projects onto $[0,\infty)$, and $K_P, K_I, K_D$ are hyperparameters for proportional, integral, and derivative control, respectively. This PID mechanism accelerates convergence to feasible policies by responsively adapting $\lambda$ to observed constraint violations.
We maintain separate value functions for reward and cost to compute Generalized Advantage Estimates $A_t^{r}$ and $A_t^{c}$, and form a Lagrangian advantage $A_t^{\lambda}=\bar{A}_t^{r}-\lambda \bar{A}_t^{c}$ with normalized advantages $\bar{A}_t^{r}$ and $\bar{A}_t^{c}$.

Standard CPPO-PID uses a symmetric clip $[1-\epsilon,\,1+\epsilon]$ to ensure conservative updates, which can slow on-hardware learning. We propose Accelerated CPPO-PID by integrating several advances. Inspired by\cite{yu2025dapo} , we conditionally enlarge only the upper clip bound when the estimated $A_t^{r}$ is positive and $A_t^{c}$ is non-positive to enhance wider exploration without sacrificing safety. Formally, with $\rho_t=\tfrac{\pi_\theta(a_t\mid s_t)}{\pi_{\theta_{\mathrm{old}}}(a_t\mid s_t)}$, we define an asymmetric upper bound:

\footnotesize 
\begin{equation}
\begin{aligned}
\epsilon_t^{+} =
\begin{cases}
\epsilon_{\mathrm{hi}}=0.28, & \text{if } A_t^{r}>0 \wedge A_t^{c}\leq0 \wedge \text{ep} \ge \text{ep}_{\mathrm{warm}},\\[2pt]
\epsilon=0.2,               & \text{otherwise},
\end{cases}
\end{aligned}
\end{equation}
\normalsize
where $\text{ep}$ is the training episode, $\text{ep}_{\mathrm{warm}}$ is a brief warm-up after which value net fits stabilize. The step-wise actor surrogate becomes:

\footnotesize 
\begin{equation}
\begin{aligned}
L_{\mathrm{step}}(\theta)=-\mathbb{E}\Bigl[\min\!\Bigl(\rho_t A_t^{\lambda},\ \operatorname{clip}\bigl(\rho_t,\,1-\epsilon,\,1+\epsilon_t^{+}\bigr)\, A_t^{\lambda}\Bigr)\Bigr].
\end{aligned}
\end{equation}

\normalsize
Locomotion episodes comprise repeated cycles. To align updates with cycle-level performance, we aggregate importance ratios over each detected cycle $p$ of length $H$.
We operate in the log domain by defining a logged signed importance ratio $\iota_t = \log \rho_t \cdot \text{sign}(A_t^{\lambda})$. Inspired by recent work on geometric aggregation\cite{zhao2025geometric}, we compute a clipped geometric mean for the cycle-wise importance ratio:

\footnotesize 
\begin{equation}
\begin{aligned}
\tilde{\rho}_p=\exp\!\Bigl[\tfrac{1}{\lvert H\rvert}\sum_{t\in p} \Bigl( \text{min}(\iota_t, \text{clip}(\iota_t, -\epsilon_p, \epsilon_p) \cdot \text{sign}(A_t^{\lambda}) \Bigr)  \Bigr].
\end{aligned}
\end{equation}

\normalsize
The geometric mean is less sensitive to outliers and provides a smoother cycle-level importance metric. To encourage exploration at this cycle level, we adopt a larger clipping value $\epsilon_p$ of up to 0.4. The corresponding cycle-level surrogate is:

\footnotesize 
\begin{equation}
\begin{aligned}
L_{\mathrm{cyc}}(\theta)=-\mathbb{E}\Bigl[\tilde{\rho}_p A_t^{\lambda}\Bigr], \ \text{where} \ \ t\in p.
\end{aligned}
\end{equation}

\normalsize
The final actor objective blends local and global views,

\footnotesize 
\begin{equation}
\begin{aligned}
L_{\mathrm{actor}}(\theta)=\alpha\, L_{\mathrm{step}}(\theta) + (1-\alpha)\, L_{\mathrm{cyc}}(\theta),
\end{aligned}
\end{equation}
\normalsize
where $\alpha\in(0,1)$ balances step-wise stability and cycle-level alignment. This hybrid objective provides a multi-faceted view that enhances both policy exploration and training stability.

Moreover, letting $\bar{A}_p$ denote the mean advantage over a cycle $\frac{1}{|H|}\sum_{t\in p}A_t^{\lambda}$, we have the actor gradient for the cycle-wise loss.

\footnotesize 
\begin{equation}
\begin{aligned}
\nabla_{\theta}L_{\mathrm{cyc}}(\theta)=-\mathbb{E}_p\!\Bigl[\bar{A}_p\,\tilde{\rho}_p\ \frac{1}{|H|}\sum_{t\in p}\mathbf{1}\{|\iota_t| \le \epsilon_p\}\, \chi_t\Bigr].
\end{aligned}
\end{equation}

\normalsize
The geometric mean aggregation not only captures the policy change over the entire cycle but is robust to outliers, filtering spikes smoothly to provide a stable, global importance metric.
Crucially, the gradient aggregation combines local gradients weighted by this stable, cycle-level statistic. Letting  
$\mathbf{g}$ denote the average gradient $\sum_{t\in p} \chi_t$, the projection of this gradient onto the average policy gradient direction within a cycle is $\nabla_{\theta}L_{\mathrm{cyc}}(\theta) \cdot \mathbf{g} = \bar{A}_p\,\tilde{\rho}_p \|\mathbf{g} \|^2$. Because $\tilde{\rho}_p > 0$ and $\|\mathbf{g} \|^2 \geq 0$, the sign of the dot product between the average gradient and the aggregated one matches the sign of $\bar{A}_p$.

This approach is effective because $\bar{A}_p$ serves as a reliable performance metric. Since the goal is to optimize sustained performance over an infinite horizon and each step’s thrust contribution is approximately uniform, we expect the discount factor $\gamma$ to approach 1. In this setting, the value function reflects the long-term expected return. The advantage $A_t$ at each step measures the immediate benefit of an action relative to this long-term average. By averaging these step-wise advantages over a full locomotion cycle to compute $\bar{A}_p$, we create a metric that filters out noisy, single-step evaluations. This cycle-averaged advantage robustly indicates whether the sequence of actions within that cycle contributed positively or negatively to the overall objective, providing a stable learning signal.

This stability provides a significant advantage for practical safe reinforcement learning. It reduces the reliance on extensive hyperparameter tuning, which is a common requirement for other methods but is often impractical for on-hardware experiments.
In our method, the Lagrange multiplier $\lambda$ increases when constraints are violated, biasing $A_t^{\lambda}$ negatively to prioritize cost reduction. When the policy persistently violates safety constraints, our cycle-level objective provides global guidance towards safer regions, avoiding the erratic updates typical of step-wise approaches.
This mitigates the tendency of CPPO-PID to temporarily sacrifice constraint satisfaction for reward, thereby reducing hyperparameter sensitivity and accelerating safe exploration.

\subsection{Policy Transferring to Quadruped}

After online safe RL training, the learned policy is transferred to the original quadrupedal swimming robot. Because the gait is periodic, the policy is first executed in the tow tank in inference mode, and one full cycle of the commanded joint trajectories is recorded as a gait primitive. This cycle is then applied to all four limbs as a baseline description of the paddle. Power and communications are provided via external cables for centralized control and data acquisition.

To preserve symmetry and reduce reaction lift fluctuation, the four limbs are partitioned into two diagonal pairs. The two pairs execute the same recorded cycle with a phase difference of half a cycle. This alignment allows peaks and troughs of pairwise forces to overlap in time, which smooths the aggregate force profile and suppresses oscillations.

The resulting gaits from different trained policies are evaluated on the quadrupedal robot. 
Each gait is tested for the farthest forward distance traveled within a fixed time window, which quantifies propulsion efficiency, oscillation suppression, and stability. 
Moreover, supplementary videos are provided for clearer and more intuitive visualization of the swimming behavior.

\section{Experiments and Results}

To evaluate the learning capability of the proposed framework and its suitability for real-world deployment, we conduct two complementary sets of experiments. First, gait optimization is performed in a towing tank, enabling controlled assessment of single-limb dynamics and constraint-aware learning. Subsequently, the optimized gaits are transferred to the quadrupedal robot and validated in free-swimming trials, demonstrating their effectiveness in achieving stable and efficient locomotion.

\subsection{Gait Optimizing}

We train the safe RL policy using towing data and on-policy rollouts. Validation trials in the towing tank assess thrust maintenance and lift reduction. The towing carriage advances at a constant speed of \(0.15\,\mathrm{m/s}\). During towing, sensor signals are Kalman-filtered and downsampled to $20\,\mathrm{Hz}$ to achieve frequency unification with the limb controller. The limb has two DoFs; for each DoF, the maximum swing amplitude is \(20^\circ\). For each training episode, we collect about 360 control steps and hyperparameters are kept consistent across experiments. The ACPPO-PID parameter \(\alpha\) is set to \(0.2\) and $\text{ep}_\text{warm}$ is set to 10.

\begin{figure}[htbp]
    \centering
    {\includegraphics[width=\linewidth]{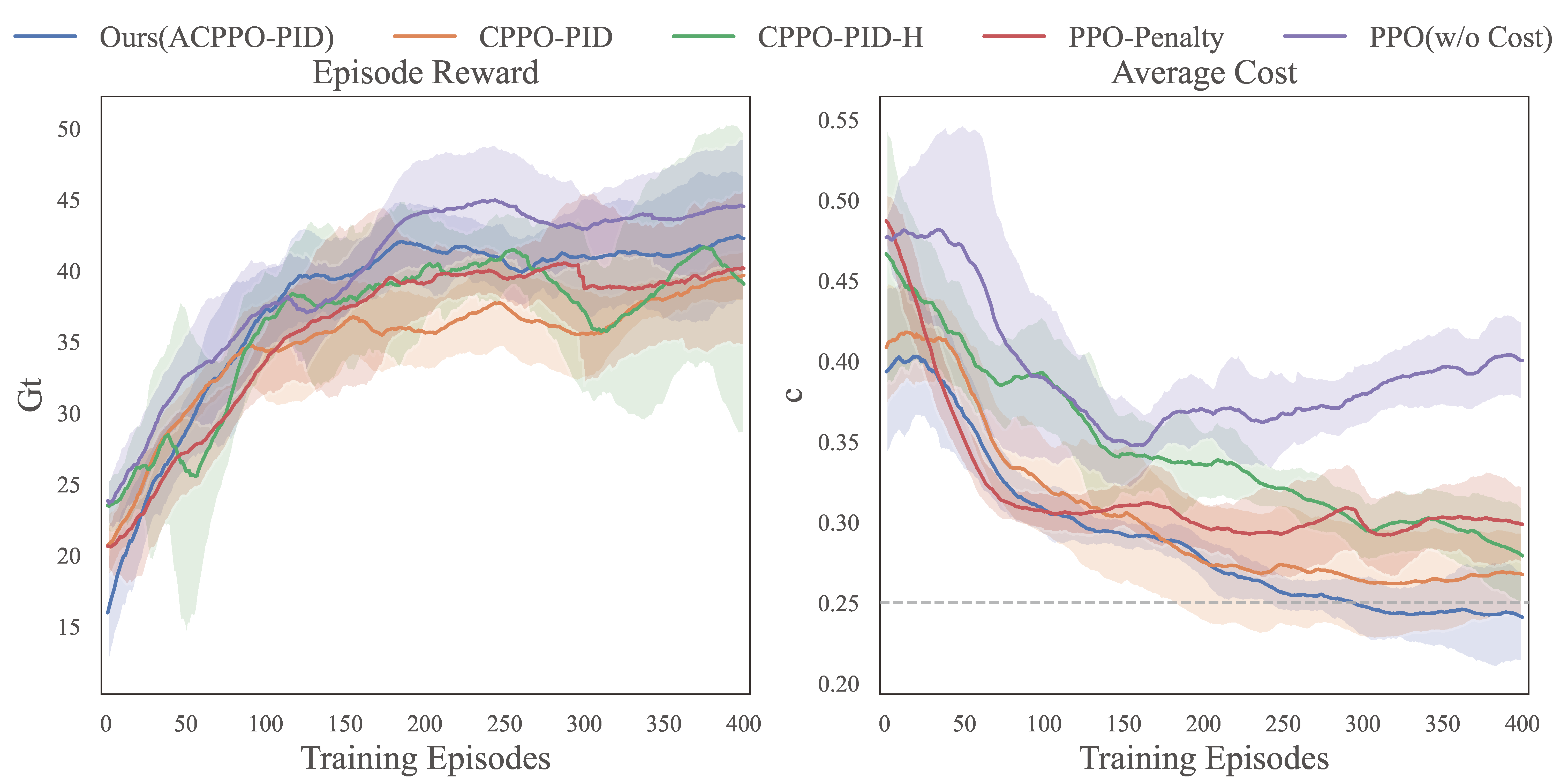}}
    \caption{
        The training curves (reward and average cost) of the baselines over 400 episodes, each method is trained in three random seeds. The gray line marks the cost limit. 
    }
    \label{fig:safe_baseline}
    \vspace{-0.05in}
\end{figure}

\subsubsection{Baselines}
We compare Our method(ACPPO-PID) against the following to demonstrate benefits in both constraint satisfaction and learning efficiency:
\begin{itemize}
    \item CPPO-PID: a SOTA safe RL algorithm, 
    \item CPPO-PID-H: a naive high-clip variant of CPPO-PID using a fixed clip $[1-\epsilon,\,1+\epsilon_\text{hi}]$ to encourage exploration irrespective of safety.
    \item PPO-Penalty: instead of optimizing a constrained objective, incorporates cost as a scalar penalty on reward $r \leftarrow r - 0.5 \cdot c$. 
    \item PPO without cost: ignores the cost signal and trains a standard PPO policy without constraints.
\end{itemize}
All methods are trained from a policy initialized by IL.


\begin{figure}[htbp]
    \centering
    {\includegraphics[width=\linewidth]{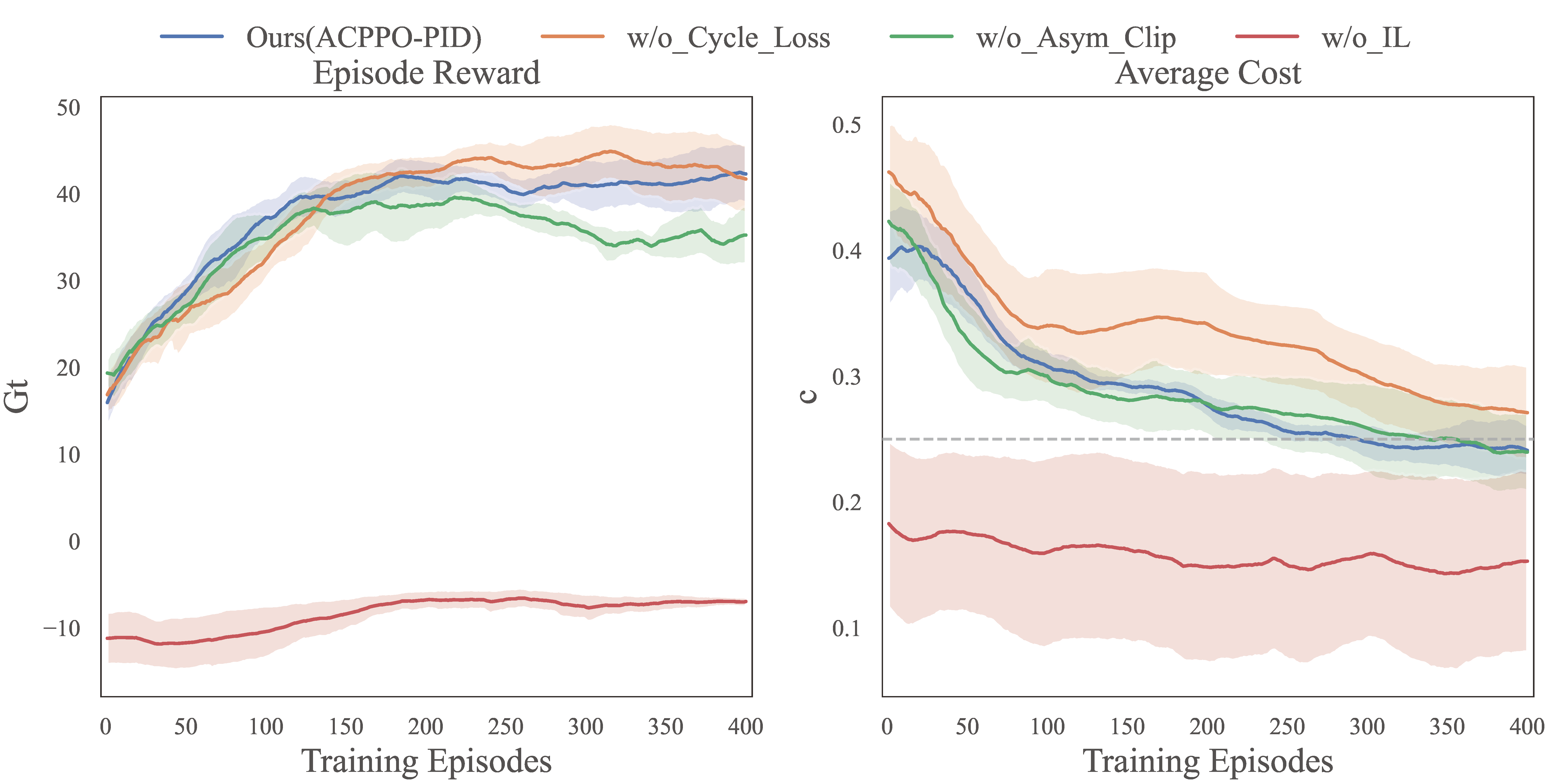}}
    \caption{
        Ablations removing the cycle‑level objective, conditional high clipping, or imitation learning quantify each component’s contribution to stability and data efficiency. Each ablation is trained in three random seeds. The gray line marks the cost limit. 
    }
    \label{fig:safe_ablation}
    \vspace{-0.15in}
\end{figure}

\subsubsection{Ablation Study}
Our method accelerates learning in three respects. To assess their contributions, we conducted ablations by removing each component in turn and compare the resulting variants:

\begin{itemize}
    \item without cycle loss: removes the cycle‑aligned objective $L_{\mathrm{cyc}}$, training solely with the step‑wise surrogate.
    \item without asymmetric clipping: enforces symmetric clipping $[1-\epsilon,\,1+\epsilon]$ at all steps, disabling the conditional upper‑bound enlargement used for safe, wider exploration.
    \item without IL: omits the imitation learning initialization and trains the policy from scratch on-hardware.
\end{itemize}



\subsubsection{Metrics}

We evaluate all methods by undiscounted episode reward and average cost (Table~\ref{table:tank_eval}). For each method, we train three independent models with different random seeds. After 400 training episodes, each model is evaluated in inference mode over three rollouts; we report the mean and standard deviation across nine runs per method. The policies trained from scratch fail to converge to a feasible motion and is excluded from the final comparison. We also include a sinusoidal paddling policy with the best parameters found by BF (denoted BF), evaluated over three rollouts.

\begin{table}[ht]
\caption{Single limb evaluated in a tow tank}
\centering
\begin{tabular}{|c|c|c|}
\hline
Algorithm & Episode Reward & Average Cost \\
\hline
CPPO-PID & \num{39.58 \pm 3.93} & \num{0.249 \pm 0.021}\\
CPPO-PID-H & \num{37.10 \pm 10.84} & \num{0.261 \pm 0.023 }  \\
PPO-Penalty & \num{43.46 \pm 3.91 } & \num{0.314 \pm 0.021 }\\
PPO (w/o Cost)  & \num{ 48.58 \pm 5.15} & \num{ 0.391 \pm 0.039}\\
\hline
Ours (ACPPO-PID) & \num{  45.19 \pm 3.83} & \num{ 0.235 \pm 0.019}\\
\hline
Ours (w/o Cycle Loss) & \num{ 43.35 \pm 5.92} & \num{ 0.271 \pm 0.027 }\\
Ours (w/o Asym Clip) & \num{38.66 \pm 5.56 } & \num{ 0.212 \pm 0.026} \\
\hline
BF & \num{26.31 \pm 1.02} & \num{0.332 \pm 0.008} \\
\hline
\end{tabular}
\label{table:tank_eval}
\vspace{-0.1in}
\end{table}

\subsubsection{Advantages of ACPPO-PID}
\textbf{Accelerated Learning.}
Fig~\ref{fig:safe_baseline} presents the learning curves of episodic reward and average cost. Table.\ref{table:tank_eval} summarizes the performance of three random seeds evaluated three times each. PPO without cost attains the highest reward but incurs severe constraint violations. Among cost-aware algorithms, ACPPO-PID achieves the best reward-cost trade-off within 400 training episodes. As an accelerated and stable advance of CPPO-PID, it attains the highest reward while maintaining the lower cost. 

Comparing with a naive variant of encouraging exploration by simply widening the clipping range to $[1-\epsilon,\,1+\epsilon_{\text{hi}}]$, denoted as CPPO-PID-H, we found it suffered from training instability.
As the aggregated advantage $A_t^{\lambda}$ is computed from the normalized reward and cost advantages, it only ranks step-wise relative advantage within a training batch. In contrast, our stepwise selection applies stricter filtering so that only genuinely advantageous updates proceed, and the cycle-wise smoothing over an entire paddle further promotes safe exploration.
Moreover, as shown in Fig.~\ref{fig:safe_ablation}, it is evident that imitation learning is essential. Without an initial policy, 400 on-hardware episodes are insufficient for the limb to discover a feasible paddling motion. 

\textbf{Stability and Constraint Satisfaction.}
As shown in Fig.~\ref{fig:safe_baseline} and Fig.~\ref{fig:safe_ablation}, the proposed method achieves consistently higher stability, whereas other approaches with larger clipping ranges exhibit pronounced fluctuations. Moreover, our method effectively suppresses cost while sustaining high performance. This advantage stems from the dual-component design of ACPPO-PID. When safety constraints are repeatedly violated, the cycle-level objective steers the entire policy update toward constraint satisfaction. In parallel, the step-wise mechanism more strictly identifies safe and advantageous actions, applying a larger update ratio to reinforce them. The synergy of these two mechanisms enables safe yet accelerated exploration, thereby validating the effectiveness of the proposed improvements.

\subsection{Quadruped Swimming}
For quantitative validation, free-swimming experiments are conducted in a $4.0\,\mathrm{m} \times 1.5\,\mathrm{m}$ water tank. The robot is tested under various gaits, each running for $6\,\mathrm{s}$ and repeated three times. Swimming distance serves as the efficiency metric: a longer average distance indicates superior time-averaged thrust and reduced ineffective work during the swing phase, thereby indirectly validating the proposed single-leg optimization method.

\begin{figure}[htbp]
    \centering
    {\includegraphics[width=0.9\linewidth]{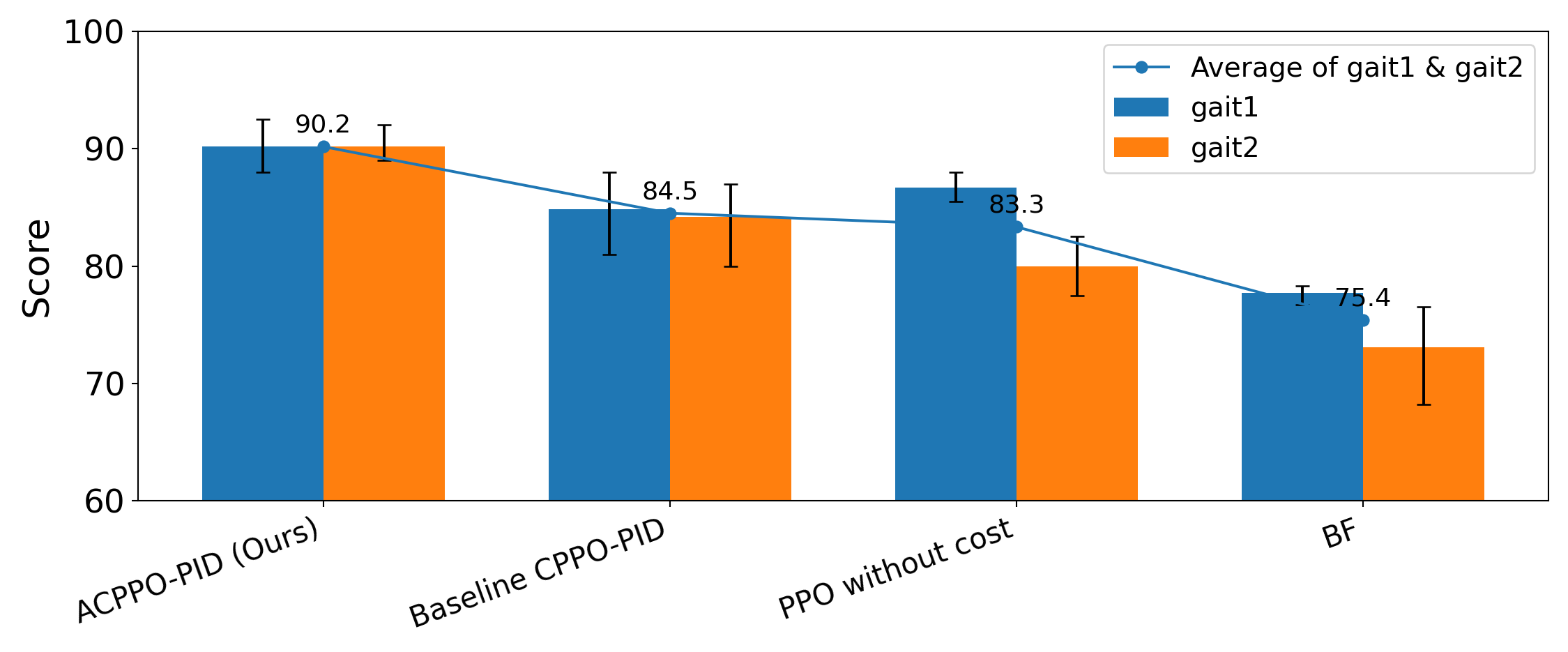}}
    \caption{
        Forward swimming performance under two representative gaits per algorithm, averaged over three trials. ACPPO-PID achieves the highest scores with lower variance, while CPPO-PID, PPO, and BF show reduced thrust and greater sensitivity to gait differences. 
    }
    \label{fig:quadruped_bar}
    \vspace{-0.15in}
\end{figure}
As shown in Fig.~\ref{fig:quadruped_bar}, we deploy two representative gaits per algorithm on the quadruped robot, each executed three times and averaged. The proposed \textbf{ACPPO-PID} achieves the best performance, outperforming the constrained baseline \textit{CPPO-PID} by about 7\%, the unconstrained \textit{PPO} by 8--9\%, and the brute-force search (\textit{BF}) by over 19--20\%. These results confirm that ACPPO-PID not only produces stronger forward thrust but also maintains more consistent performance across gaits, achieving a better balance of propulsion efficiency and stability.

\begin{figure}[htbp]
    \centering
    {\includegraphics[width=0.9\linewidth]{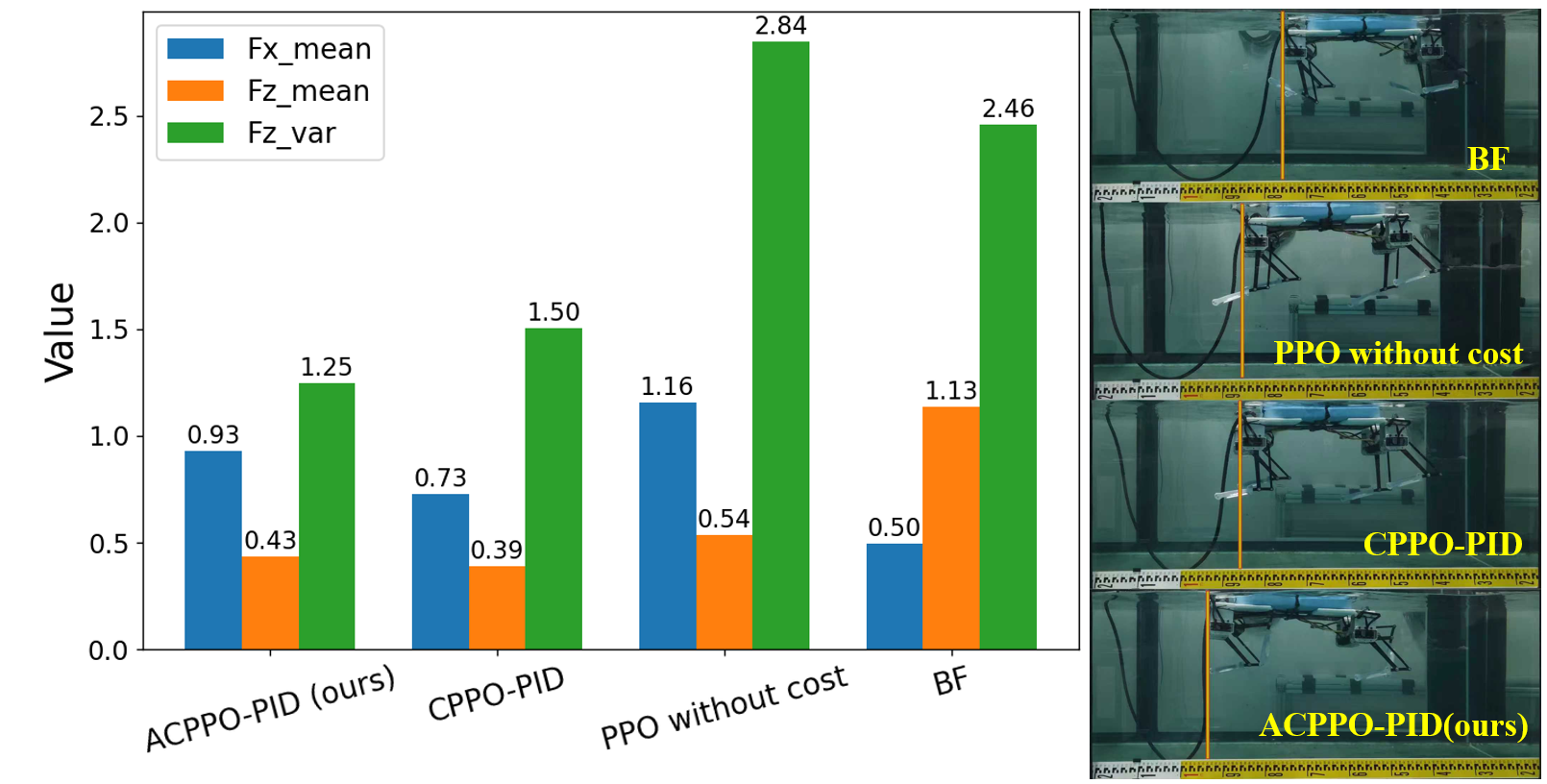}}
    \caption{
Comparison of $F_{x}^{\text{mean}}$, $F_{z}^{\text{mean}}$, and $F_{z}^{\text{var}}$ across algorithms, showing ACPPO-PID achieves strong thrust with reduced lift fluctuations.The right panel illustrates the \textbf{maximum} displacement achieved in this experiment under parameterized motion, standard PPO training, and our proposed ACPPO-PID.
    }
    \label{fig:quadruped_force}
    \vspace{-0.1in}
\end{figure}

It is worth noting that for the \textit{PPO without cost} algorithm, gait 1 shows relatively good displacement performance. To better understand this phenomenon, we further analyze the net forces and lift generated by the bio-inspired propulsors under the best-performing gait of each algorithm, computing the mean forward thrust ($F_{x}^{\text{mean}}$), mean lift ($F_{z}^{\text{mean}}$), and lift variance ($F_{z}^{\text{var}}$) over time. The results in Fig.~\ref{fig:quadruped_force} reveal clear quantitative differences. \textbf{ACPPO-PID (ours)} achieves a mean forward thrust of $0.93$, which is approximately $27\%$ higher than \textit{CPPO-PID} and $86\%$ higher than \textit{BF}, while being only $24.7\%$ lower than unconstrained \textit{PPO}. More importantly, ACPPO-PID effectively suppresses destabilizing lift: its mean lift is $20.4\%$ smaller than PPO and $61.9\%$ smaller than BF, with only a slight increase of $10.3\%$ compared with CPPO-PID. In terms of stability, ACPPO-PID shows the lowest lift variance among strong performers, reducing it by $55.9\%$ relative to PPO, $49.2\%$ relative to BF, and $16.7\%$ relative to CPPO-PID. These results indicate that ACPPO-PID strikes the best trade-off in retaining strong propulsion while significantly reducing destabilizing forces. This allows the robot to maintain a more stable vertical position and thereby achieve greater net displacement.

\begin{figure}[t]
    \centering
    {\includegraphics[width=1.0\linewidth]{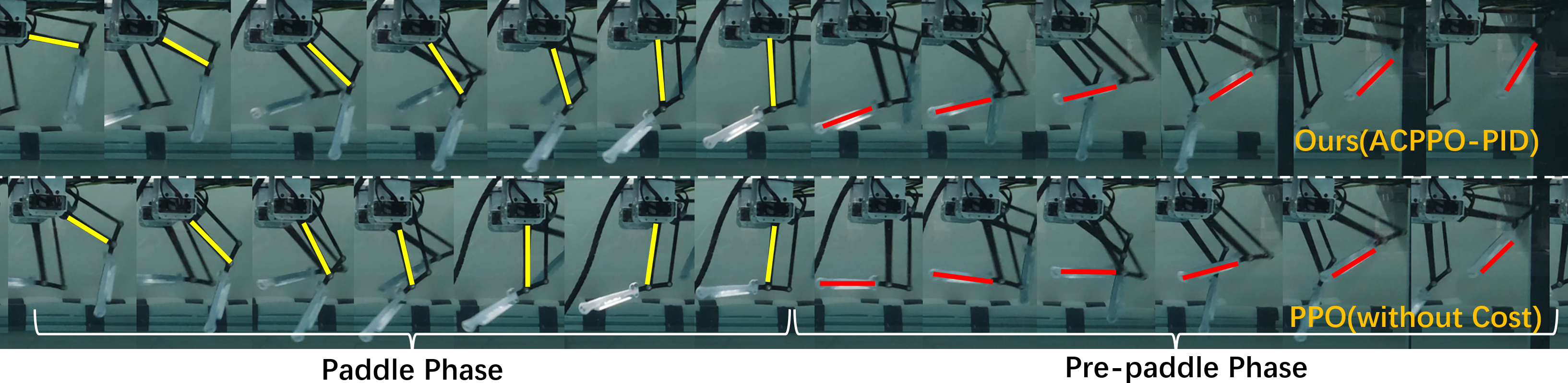}} 
    \caption{Comparison of limb motion patterns between our constrained method (ACPPO-PID, top) and unconstrained PPO (bottom) during one locomotion cycle. Key differences are highlighted: calf movement during pre-paddle phase (red) and hip movement during paddle phase (yellow). }
    \label{fig:gesture}
    \vspace{-0.15in}
\end{figure}

\textbf{A closer look at optimized gaits.} Fig~\ref{fig:gesture} compares limb motions between our method (top) and unconstrained PPO (bottom). Each cycle is divided into pre-paddle and paddle phases based on thrust generation, with key differences highlighted in calf movement (red) and hip movement (yellow).
During the pre-paddle phase, our method adjusts the calf to maintain velocity direction parallel to calf orientation to reduce drag, while PPO fixes a small angle between calf and hip, positioning for greater extension and larger thrust-generating surface area in the subsequent phase.
During the paddle phase, our method drives the hip backward directly, whereas PPO tilts forward then back. This creates a larger range of motion that enhances thrust generation but introduces significant additional lift.

Overall, PPO aggressively pursues thrust at the cost of lift-induced instability, while our constrained approach produces cleaner, more stable, and thrust-centric motion patterns that better achieve optimization objectives.

\section{Conclusion}
This work demonstrates that explicit constraints on lateral forces and oscillatory amplitudes are essential for stable and efficient quadruped swimming. By formulating gait optimization as a constrained reinforcement learning problem, we enhance propulsion efficiency under strong fluid–structure interactions without sacrificing stability. To this end, we propose \textbf{ACPPO-PID}, a safe RL framework for oscillatory propulsion that selectively widens exploration only when advantages are safe and positive and regulates updates with cycle-level performance. Experiments on both single actuators and quadruped robots confirm that ACPPO-PID consistently outperforms strong baselines on the reward–cost frontier within limited hardware budgets. 
Current validation was limited to quiescent water, leaving performance under extreme hydrodynamic disturbances, such as strong currents or turbulence, as an open challenge. Future work will therefore focus on integrating online adaptation and domain randomization mechanisms to mitigate these environmental uncertainties and ensure robust deployment in open-water environments.

\bibliographystyle{IEEEtran}

\bibliography{main}  

\end{document}